%% file: iclr2026_conference.tex
\documentclass{article} % For LaTeX2e
\usepackage{iclr2026_conference,times}

\input{math_commands.tex}

\usepackage{hyperref}
\usepackage{url}
\usepackage{array}
\usepackage{boldline}
\usepackage{subscript}
\usepackage[table]{xcolor}
\usepackage{booktabs}

\usepackage{multirow}
\usepackage{amsmath}
\usepackage{graphicx}
\usepackage{algorithm}
\usepackage{algpseudocode}
\usepackage{wrapfig}
\usepackage{amssymb}

\newtheorem{definition}{Definition}
\newtheorem{proposition}{Proposition}

\definecolor{deepgreen}{rgb}{0,0.5,0} 

\newcommand{\methodname}{DCFold~}
\newcommand{\methodnamecomma}{DCFold}

\title{DCFold: Efficient Protein Structure Generation with Single Forward Pass}

\author{Zhe Zhang$^{1,2}$ \thanks{Zhe Zhang (zhe-zhan22@mails.tsinghua.edu.cn).} \quad Yuanning Feng$^{1,3}$ \quad
Yuxuan Song$^{1,4}$ \quad Keyue Qiu$^1$ \quad
Hao Zhou$^{1}$ \thanks{Hao Zhou
(zhouhao@air.tsinghua.edu.cn).}%
\quad Wei-Ying Ma$^1$\\
$^1$ Institute for AI Industry Research (AIR), Tsinghua University \\
$^2$ Department of Computer Science and Technology, Tsinghua University \\
$^3$ School of Computer Science and Technology, Huazhong University of Science and Technology \\
$^4$ ByteDance Seed
}

\iclrfinalcopy % Uncomment for camera-ready version, but NOT for submission.
\begin{document}

\maketitle

\begin{abstract}
AlphaFold3 introduces a diffusion-based architecture that elevates protein structure prediction to all-atom resolution with improved accuracy. 
% The strength of this performance has established AlphaFold3 as a solid and widely adopted foundation model for diverse generation and design tasks. 
This state-of-the-art performance has established AlphaFold3 as a foundation model for diverse generation and design tasks.
However, its iterative design substantially increases inference time, limiting practical deployment in downstream settings such as virtual screening and protein design. We propose \methodnamecomma, a single-step generative model that attains AlphaFold3-level accuracy. 
% Through our Dual Consistency training framework, anchored by the key component Temporal Geodesic Matching (TGM), \methodname achieves a \textbf{15×} acceleration in inference while maintaining predictive fidelity. 
Our Dual Consistency training framework, which incorporates a novel Temporal Geodesic Matching (TGM) scheduler, enables \methodname to achieve a \textbf{15×} acceleration in inference while maintaining predictive fidelity.
We validate its effectiveness across both structure prediction and binder design benchmarks.

\end{abstract}

\section{Introduction}
\begin{wrapfigure}{r}{0.55\textwidth}
    \label{fig:time}
    \centering
    \includegraphics[width=0.55\textwidth]{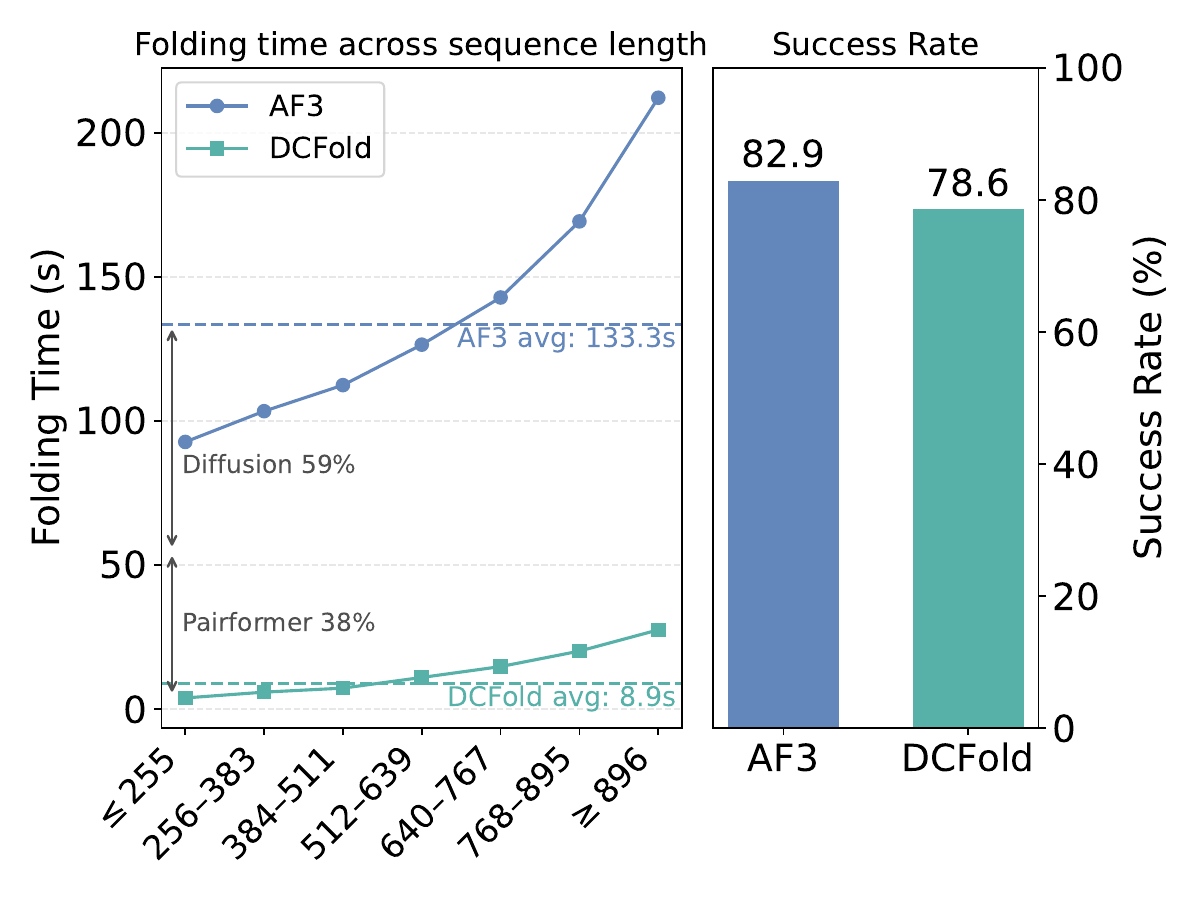}
    \caption{The acceleration ratio and generative quality of \methodname on Posebusters V2.}
\end{wrapfigure}

Proteins realize their biological functions through intricate three-dimensional conformations, and predicting such structures has long been a central challenge in computational biology. AlphaFold2 marked a breakthrough by combining multiple sequence alignments with geometric constraints in an end-to-end framework, achieving near-experimental accuracy \citep{jumper2021highly}. Building on this foundation, AlphaFold3 reformulates the architecture into an all-atom framework and introduces a diffusion-based structure module, thereby enabling the generative modeling of not only proteins but also a wide spectrum of biomolecular complexes \citep{abramson2024accurate}. Consequently, this series of models are widely adopted as foundation models for downstream applications such as virtual screening and protein design  \citep{alhumaid2024reliability, baselious2024comparative, jendrusch2025alphadesign, frank2024scalable, bennett2023improving}. However, AlphaFold3's architecture, which relies on iterative Pairformer recycling and multi-step diffusion \citep{ho2020denoising}, requires substantially greater computational overhead than AlphaFold2, restricting its accessibility in downstream workflows.

More specifically, we observe that on long sequences, the execution time of AlphaFold3 is measured in minutes, which severely limits its usability in downstream tasks that demand high throughput. For instance, small-scale laboratory screening often requires predictions for thousands of candidates \citep{li2023machine}, and when extended to large public databases, this number grows to an infeasible scale; protein design tasks typically involve comparable computational demand. Previous work such as BindCraft has attempted to mitigate this by manually reducing the number of recycling iterations on simpler structures, thus trading accuracy for efficiency \citep{pacesa2024bindcraft}. However, such compromises inevitably degrade predictive performance. Moreover, in hallucination-based approaches, the multistep iterative refinement process hinders feasible gradient backpropagation, ultimately preventing the broader community from adopting AlphaFold3 as a foundation model for diverse applications.

To accelerate the diffusion process, recent advances in generative modeling have explored the use of high-order solvers and consistency models. While high-order solvers improve efficiency, they rarely reduce the number of sampling steps below 10 \citep{lu2022dpm, zhao2023unipc}. Consistency models, on the other hand, have achieved remarkable success in image generation and benefited from refined training schedules \citep{song2023consistency, song2023improved, lu2024simplifying}. However, directly applying them to AlphaFold3 faces two major challenges: (i) standard schedules assume fixed-dimensional data and pair steps by a constant Euclidean distance, which fails to accommodate variable protein sequence lengths and leads to unstable training dynamics (details in Section~\ref{sec:TGM_exp}); and (ii) AlphaFold3’s architecture also relies on iterative Pairformer recycles, introducing an additional bottleneck that conventional diffusion consistency methods cannot address. 

To address these challenges, we propose \methodnamecomma, a single-step folding model trained under Dual Consistency framework that attains AlphaFold3-level accuracy. We mitigate the inference bottleneck by jointly enforcing Pairformer Consistency and Diffusion Consistency, thereby eliminating both sources of iterative overhead. Crucially, we address the fundamental challenge of diffusion acceleration through rigorous theoretical derivations, and subsequently introduce a novel Temporal Geodesic Matching (TGM) scheduler, which adaptively pairs timesteps in the intrinsic geometric space of proteins. Together, these innovations preserve the predictive accuracy of AlphaFold3 while drastically reducing inference costs, enabling one-step predictions that are both efficient and reliable.

We extensively validate the effectiveness of \methodname on structure prediction benchmarks, which provide a rigorous and fair evaluation protocol. Beyond this standard setting, we further assess \methodname in the more practical binder design tasks, where both inference speed and structural accuracy are critical to this setting.

In short, we summarize our contributions as follows:

\begin{itemize}
    \item We propose \methodnamecomma, an inference-efficient structure prediction model that achieves performance and flexibility comparable to state-of-the-art applications. By leveraging the Dual Consistency framework, \methodname eliminates the iterative overhead inherent in AlphaFold3’s architecture.
    \item We identify the key limitations of conventional consistency model (CM) methods when applied to variable-length protein sequences, and introduce Temporal Geodesic Matching (TGM) for a novel consistency schedule that both stabilizes training and yields improved performance.
    \item We evaluate the performance of \methodname across a diverse set of benchmarks and settings. On both Posebusters V2 and Recent PDB, it reaches AlphaFold3-level accuracy while achieving a notable $15\times$ speedup. Implemented in the binder design pipeline, \methodname demonstrates strong foundational capabilities while employing a lightweight architecture that ensures feasible gradient propagation. This design significantly improves the success rate of in silico screening by enabling faster and more reliable candidate evaluation.
\end{itemize}

\section{Preliminary}
Diffusion models have emerged as a powerful class of generative models, achieving state-of-the-art performance across image, audio, and molecular generation tasks \citep{ho2020denoising, rombach2022high, trippe2022diffusion}. A key limitation of standard diffusion samplers is their reliance on dozens to hundreds of function evaluations, which renders inference prohibitively expensive in high-dimensional settings such as protein folding. To address this bottleneck, recent work has focused on diffusion acceleration, aiming to distill or redesign the sampling process into far fewer steps. Among these approaches, \emph{Consistency Models} (CMs) \citep{song2023consistency} provide a principled framework built upon the probability flow ODE (PF-ODE), which establishes a bijective mapping between the clean data distribution and the noise distribution. CMs introduce a consistency function $f_\theta(x_t, t)$ that directly maps a noisy sample $x_t$ at time $t$ back to the clean signal $x_0$, subject to the boundary condition $f_\theta(x_0,0)=x_0$. Training then proceeds by discretizing the PF-ODE into a curriculum of time intervals ${t_i}$, and minimizing a loss that enforces functional consistency across adjacent timesteps,
\begin{equation}
    \mathcal{L}_{\text{CM}} = \E\left[w(t_i)d\left(f_\theta(x_{t_{i+1}},t_{i+1}), f_{\theta^-}(\tilde{x}_{t_i},t_i)\right)\right],
\end{equation}

where $w: \mathbb{R}_{\ge 0} \to \mathbb{R}^+$ denotes a positive weighting function, $d(\cdot,\cdot)$ is a metric function, $\theta^-$ is an EMA copy of the network, and $\tilde{x}_{t_i}$ is obtained by one-step PF-ODE integration. This objective ensures that the model predictions are invariant to the choice of sampling timestep, thereby collapsing multi-step trajectory into a single-step or few-step generator. Building on this foundation, subsequent refinements such as iCT \citep{song2023improved}, sCM \citep{lu2024simplifying}, and ECM \citep{geng2024consistency}, have optimized the weighting functions, discretization schedules, and training methodologies, resulting in substantial improvements in both efficiency and sample quality.

\section{Method}
\begin{figure}
    \centering
    \includegraphics[width=1\linewidth]{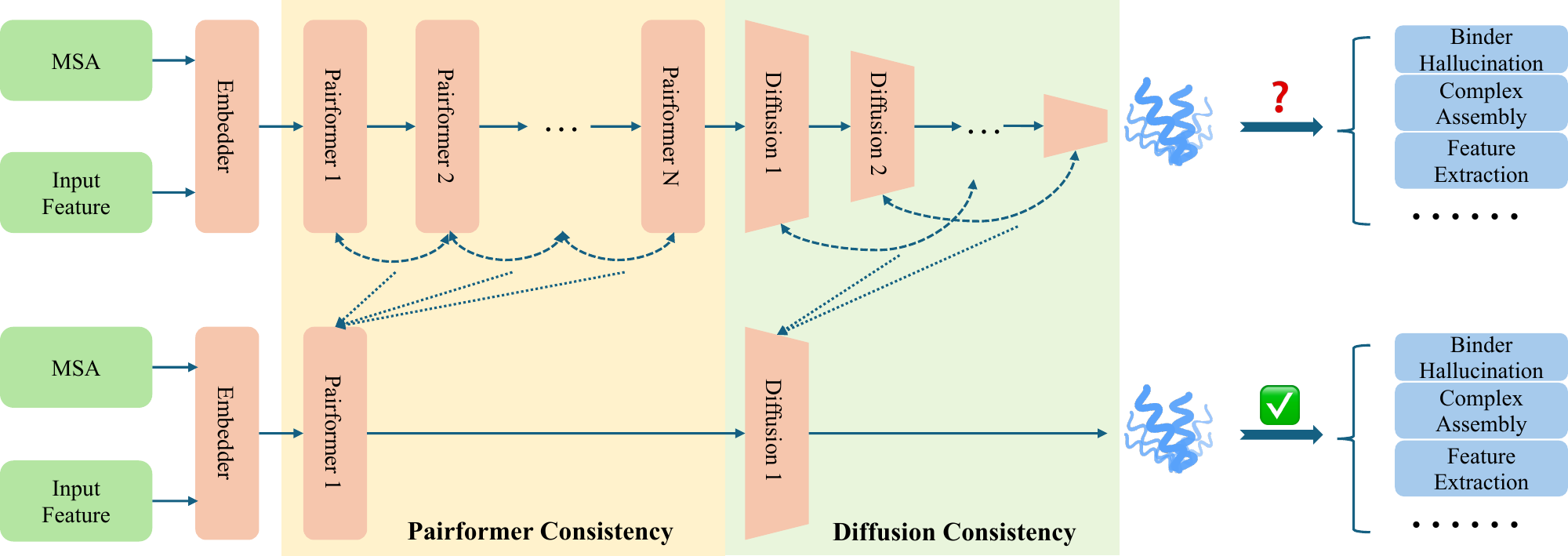}
    \caption{Overview of Dual Consistency framework (top: AlphaFold3; bottom: \methodnamecomma).}
    \label{fig:pipeline}
\end{figure}

\subsection{Overview}
We introduce \methodnamecomma, a high-accuracy single-step predictor. In Section \ref{sec:dual_consistency}, we describe the components of the Dual Consistency framework, which enforces consistency across the two major bottlenecks of AlphaFold3. In Section \ref{sec:TGM}, we zoom in on the diffusion acceleration challenge and identify the key issue with prior consistency-based methods when training on variable-length sequences within diffusions. To tackle this challenge for complex structure prediction, we propose Temporal Geodesic Matching (TGM), which stabilizes training on the protein sequence modality.

\subsection{Dual Consistency}
\label{sec:dual_consistency}
\begin{wraptable}{r}{0.55\textwidth}
\centering
\resizebox{\linewidth}{!}{
\begin{tabular}{ccccc}
\toprule
\textbf{Stage} & \textbf{Module} & $\mathcal{L}_{\text{confidence}}$ & $\mathcal{L}_{\text{diffusion}}$ & $\mathcal{L}_{\text{pairformer}}$ \\
\hline
(i) & Diffusion & $10^{-4}$ & 1 & $\times$ \\
(ii) & Pairformer & $10^{-4}$ & $\times$ & 1 \\
\bottomrule
\end{tabular}
}
\caption{Training stages and the weights of each term.}
\label{tab:two_stage}  
\end{wraptable}

We identify the major factors impeding AlphaFold3’s inference efficiency as the iterative diffusion process and Pairformer recycling, as illustrated in Figure~\ref{fig:time}. To address the first challenge, we investigate the behavior of AlphaFold3 under few-step sampling and find that its failure primarily arises from the sampling procedure itself. The default strategy of injecting extra stochastic noise and enlarging the ODE step size turns out to be detrimental in this regime: the enlarged step size significantly amplifies the bias in ODE predictions. To stabilize performance, we modify the sampler by disabling noise injection (setting the noise factor $\gamma_0=0$), fixing the rescaling factor $\lambda=1$, and normalizing the step size with $\eta=1$, thereby enabling stable one-step sampling.

The first challenge concerns computational efficiency. After enabling one-step sampling, the Pairformer becomes the critical bottleneck. To tackle this, we introduce \textbf{Dual Consistency}, which applies consistency learning to both the diffusion module and the Pairformer.

\paragraph{Diffusion Consistency} Although we already have a functional one-step sampler, we aim to maximize its utility. Specifically, we apply consistency distillation to the diffusion module, aligning its single-step performance with that of the multi-step counterpart, which also provides a natural warm-up for the subsequent Pairformer consistency stage. The training objective minimizes the MSE between the outputs of the diffusion module at timestep $t$ and a reference timestep $r$. Formally, the diffusion consistency loss is
\begin{equation}
    \mathcal{L}_{\text{diffusion}}=\mathbb{E}_{x,t,r,\epsilon}\left[w(t) \text{MSE}\left(f_\theta(x_t,t) - f_{\text{sg}(\theta)}(x_r,r)\right) \right],
\end{equation}

where $f_\theta$ denotes diffusion module parameterized by $\theta$, and $\text{sg}(\theta)$ denotes ``stop-gradient" operator. We find $w(t)$ to have negligible effect in experiments and therefore set $w(t)=1$.

\paragraph{Pairformer Consistency} For the most critical bottleneck in AlphaFold3, Pairformer, we observe that the architecture updates internal protein representations iteratively across multiple cycles. While increasing the number of cycles generally improves prediction accuracy, it also scales inference time linearly. Importantly, because each Pairformer cycle depends on the output of the previous one, a single forward pass through the network inherently provides representations corresponding to different cycle depths. This allows us to assess the model’s progressive refinement of structural accuracy without the need for explicit time sampling as required in diffusion-based denoising processes.

To exploit this property, we introduce a \textbf{cycle consistency loss}. Suppose pairformer runs for $N$ cycles (with $N=4$ in our experiments). After the $n$-th cycle, the model produces a pair representation $z_n$ and a single representation $s_n$. We directly adopt the \textit{total transmission error} as the loss function:
\begin{equation}
    \mathcal{L}_{\text{pairformer}} = \sum_{i=1}^{N-1} \left( \text{MSE}\left( z_i, z_{i+1} \right) + \text{MSE}\left( s_i , s_{i+1} \right) \right).
\end{equation}

Notably, we adopt the weighting strategy from AlphaFold’s supervised MSE loss. In particular, positions corresponding to nucleic acids and small molecules are assigned higher weights than amino acids. This ensures that structurally sensitive residues contribute proportionally to the loss. Let the column vector $\boldsymbol{\alpha}$ denote the per-token weighting coefficient used in AlphaFold3. For the single representations in both Diffusion Consistency and Pairformer Consistency, we directly apply $\boldsymbol{\alpha}$ as the weight. In contrast, for the pair representations in Pairformer, we adopt a multiplicative composition, using $\sqrt{\boldsymbol{\alpha}} \sqrt{\boldsymbol{\alpha}}^\top$ as the weighting matrix, where the square root is applied element-wise.

We further find that incorporating the confidence loss $\mathcal{L}_{\text{confidence}}$ from AlphaFold3’s confidence head improves training stability, where $\mathcal{L}_{\text{confidence}}$ is defined as:
\[
\mathcal{L}_{\text{confidence}} = \mathcal{L}_{\text{plddt}} + \mathcal{L}_{\text{pde}} + \mathcal{L}_{\text{resolved}} + \alpha_{\text{pae}} \cdot \mathcal{L}_{\text{pae}},
\]

where $\alpha_{\text{pae}} = 1$, and the definitions of all loss terms follow AlphaFold3. Consequently, our training procedure can be summarized in two stages: (i) train a one-step sampler, where only the diffusion module is updated, with the training objective given by $\mathcal{L}_{\text{confidence}}$ and $\mathcal{L}_{\text{diffusion}}$; (ii) apply pairformer consistency, where only a 16-block Pairformer is updated, with the training objective given by $\mathcal{L}_{\text{confidence}}$ and $\mathcal{L}_{\text{pairformer}}$. We summarize the weights of them in Table \ref{tab:two_stage}.

\begin{algorithm}[t]
\caption{Temporal Geodesic Matching (TGM)}
\begin{algorithmic}[1]
\Require Dataset $\mathcal{D}$, pretrained diffusion model $\theta$, noise distribution $p(t)$, weighting function $w(t)$, training progress $u = \frac{\text{steps}}{\text{max\_steps}} \in [0, 1]$
\While{$\theta$ not converged}
    \State Sample $x_0 \sim \mathcal{D}$, $\epsilon \sim \mathcal{N}(0, I)$, $t \sim p(t)$
    \State $r' \gets \max\left(r(t, u), 0\right)$
    \State $x_t \gets x_0 + t\epsilon$; \quad $x_{r'} \gets x_0 + r'\epsilon$
    \State $\mathcal{L} \gets w(t) \,\Vert f_\theta(x_t,t) - f_{\text{sg}(\theta)}(x_{r'},r') \Vert_2^2$ \Comment{using the same random seed}
    \State $\theta \gets \theta - \eta \nabla_\theta \mathcal{L}$
\EndWhile
\end{algorithmic}
\label{alg:TGM}
\end{algorithm}

\subsection{Temporal Geodesic Matching}
\label{sec:TGM}
While consistency-based methods have shown promise, directly applying them to complex architectures like AlphaFold often results in weight collapse, high training cost, or reliance on task-specific mappings. The core issue lies in scheduling for variable-size outputs such as protein structures. Conventional schedulers pair timesteps $(t,r)$ at fixed Euclidean intervals, producing an ill-posed curriculum: on long sequences, even small $\Delta t$ triggers drastic distribution shifts that demand unrealistic predictive leaps, whereas on short sequences the same interval provides only weak signals. This mismatch overlooks the non-uniform accumulation of information with data dimensionality, leading to instability and collapse.

To address these limitations, we introduce Temporal Geodesic Matching (TGM), a general and scalable distillation framework. TGM explicitly selects training pairs $(t, r)$ such that their geodesic distance on the temporal information manifold is preserved, thereby offering a principled mechanism to stabilize training and extend consistency learning to large-scale protein modeling tasks. Unlike Euclidean-based heuristics, TGM aligns the distillation dynamics with the intrinsic statistical geometry of the diffusion trajectory. By doing so, it ensures stability and fidelity even in high-dimensional structured output spaces such as protein backbones.

We begin by formalizing the diffusion trajectory as a geometric object. Let ${p_t(x)}_{t\in[0,T]}$ denote the family of intermediate distributions induced by the forward diffusion process. We interpret it as a coordinate charting a one-dimensional \textbf{temporal information manifold} $\mathcal{M}_t$, where each point corresponds to a distribution $p_t(x)$.

\begin{definition}
We define the temporal metric via the Fisher information with respect to the diffusion time $t$, which we refer to as the \textbf{temporal Fisher information}, and use it as the Riemannian metric tensor of $\mathcal{M}_t$:
\begin{equation}
    g(t) := \mathcal{I}(t) = \E_{p_t(x)} \left[ \left( \frac{\partial}{\partial t} \log p_t(x)\right)^2\right].
\end{equation}
\end{definition}

\begin{definition}
On the manifold where the temporal Fisher information serves as the Riemannian metric tensor, the \textbf{geodesic distance} between two time points $t$ and $r$ is defined as the corresponding geodesic length:
\begin{equation}
    d_g(t,r)=\int_r^t\sqrt{\mathcal{I}(\tau)} d\tau.
\end{equation}
\end{definition}

Our central thesis is that a stable and efficient distillation process must be grounded in the Kullback-Leibler (KL) divergence, as this is the canonical metric underlying the variational objective of diffusion models. We motivate the introduction of the Fisher information through the following theorem:

\begin{proposition}
\label{prop:KL-metric_equiv}
(Local Metric-KL Equivalence) For a small step $\Delta t = t-r \geq 0$, the geodesic distance between neighboring distributions is given by:
\begin{equation}
    d_g(t,r) = \sqrt{2} D_{\mathrm{KL}}\left( p_r(x) \Vert p_t(x)\right)^{\frac12} + \mathcal{O}\left((\Delta t)^3\right).
\end{equation}
\end{proposition}

The proof of Proposition \ref{prop:KL-metric_equiv} is provided in the Appendix \ref{app:metric-kl_proof}. The metric $d_g$ provides a principled measure of distributional discrepancy along the temporal axis. Building on this, TGM stabilizes training by enforcing a consistent alignment rule: for a given training progress $u = \frac{\text{steps}}{\text{max\_steps}} \in [0,1]$, each timestep $t$ is paired with a reference point $r$ at a fixed temporal distance, i.e., $d_g(t,r) = C(u)$, where $C(u)$ is a monotonically decreasing function. In our experiments, we specify $C(0) = C_0$ as a hyperparameter, $C(1) = 0, C(u) = C_0(1-u)^\beta, \beta>0$, and approximate $r(t,u)=t-\frac{C_0}{\sqrt{\mathcal{I}(t)}} \left(1-u\right)^\beta$ via one-step Euler method. While it is also feasible to employ higher-order numerical solvers, we did not observe significant performance gains from doing so. Furthermore, we provide the analytical form of $\mathcal{I}(t)$:

\begin{proposition}
For any diffusion model that satisfies the classical setting of $p_t(x|x_0) = \mathcal{N}(x;\mu=\alpha(t)x_0, \sigma^2(t)I)$:
\begin{equation}
    \mathcal{I}(t)=\mathbb{E}_{x_0 \sim p_{\mathrm{data}}}
\left[
\left(\frac{\dot{\sigma}(t)}{\sigma(t)}\right)^2 \cdot 2D
+\left(\frac{\dot{\alpha}(t)}{\sigma(t)}\right)^2
\Vert x_0 \Vert^2
\right],
\end{equation}
where $D$ denotes the dimensionality of the vector.
\end{proposition}

This analytical form underscores the universality of TGM. In most generative tasks, data can naturally be represented as fixed-length vectors. Furthermore, when normalized (as in image generation) or invariant to random rotations (as in protein folding), the $\Vert x_0\Vert ^2$ term admits a simplification to $\text{Var}(x_0)$ under the assumption $\mathbb{E}[x_0]=0$. In our experiments, due to AlphaFold’s adoption of the EDM framework\citep{karras2022elucidating}, we present here the specific form of $\mathcal{I}(t)$ that is used:
\begin{equation}
    \mathcal{I}(t) =
\frac{
  2D \cdot p^2
  \left(s_{\text{max}}^{1/p}-s_{\text{min}}^{1/p}\right)^2
}{
  \left[
    s_{\text{max}}^{1/p}
    +(1-t)\left(
      s_{\text{min}}^{1/p}-s_{\text{max}}^{1/p}
    \right)
  \right]^2
},
\end{equation}

where the definition of $s_{\text{min}}$ and $s_{\text{max}}$ follow EDM, which are used in AlphaFold3’s diffusion process to control the noise strength. Here we incorporate the data dimensionality $D$ into the training schedule to balance the differences in learning difficulty across amino acid sequences of varying lengths. Importantly, as the dimensionality increases, the KL divergence between distributions accumulates linearly, causing classical consistency training to exaggerate information disparities for long sequences. And we provide in Algorithm \ref{alg:TGM} the procedure for applying TGM to the diffusion module.

\subsection{Downstream Task}
After ensuring the consistency of AlphaFold3, we find that our method now holds substantial potential for downstream applications. As a representative example, we validate the effectiveness of \methodname in the task of binder design. This task typically requires models to perform large-scale sampling, followed by stringent multi-stage filtering to eliminate implausible sequences, leaving only a small subset of viable candidates. Moreover, in binder hallucination–based design frameworks, the network must be fully differentiable and amenable to gradient-based optimization \citep{pacesa2024bindcraft}. These properties make \methodname particularly well-suited for this setting, allowing it to fully demonstrate its performance advantages. The experimental details are presented in Section \ref{sec:binder_exp}.

\section{Experiment}
We design our experiments to evaluate both the accuracy and practical utility of \methodnamecomma. In Section~\ref{sec:structure_exp}, we evaluate the structural prediction capability of \methodnamecomma, showing that \methodname matches or surpasses AlphaFold3 while reducing cost. In Section~\ref{sec:binder_exp}, we assess binder hallucination, demonstrating that the reshaped output distribution improves downstream design success. Section~\ref{sec:TGM_exp} isolates the effect of TGM and shows its advantage over prior consistency schedules. Together, these results highlight the efficiency, stability, and applicability of \methodname across protein modeling tasks.

\subsection{Structure Prediction}
\label{sec:structure_exp}
\begin{table}[t]
\centering
\caption{Posebusters V2 RMSD benchmark results. We report the percentage of predictions with RMSD below different thresholds.}
\label{tab:posebusters-rmsd}
\vspace{0.1in}
\begin{tabular}{lcccccccc}
\toprule
\multirow{2}{*}{Method} & \multicolumn{4}{c}{Best (\%)} & \multicolumn{4}{c}{Worst (\%)} \\
\cmidrule(lr){2-5} \cmidrule(lr){6-9}
& $<1$ & $<2$ & $<3$ & $<5$ & $<1$ & $<2$ & $<3$ & $<5$ \\
\midrule
AlphaFold3 & \textbf{67.14} & \textbf{82.86} & \textbf{87.14} & 93.81 & 45.71 & 70.00 & 79.05 & 87.62 \\
AF3 ODE & 51.43 & 74.77 & 83.81 & 92.38 & 37.62 & 66.19 & 75.71 & 87.62 \\
\rowcolor{gray!15} \methodname (Ours) & 58.10 & 78.57 & 86.67 & \textbf{94.29} & \textbf{46.67} & \textbf{71.43} & \textbf{80.00} & \textbf{90.48} \\
\bottomrule
\end{tabular}
\end{table}

\begin{table}[t]
\centering
\caption{TM-score and Success Rate (SR) on different protein categories in the Homology Recent PDB dataset. Values in parentheses denote the absolute improvement relative to AF3 ODE.}
\label{tab:tm_sr_comparison}
\vspace{0.1in}
\resizebox{\linewidth}{!}{
\begin{tabular}{lcccccc}
\toprule
 & \multicolumn{2}{c}{PL-complex} & \multicolumn{2}{c}{Monomer} & \multicolumn{2}{c}{PP-complex} \\
\cmidrule(lr){2-3} \cmidrule(lr){4-5} \cmidrule(lr){6-7}
Method & TM-score & SR (\%) & TM-score & SR (\%) & TM-score & SR (\%) \\
\midrule
AF3 ODE 
  & 0.815 & 92.3 & 0.830 & 92.9 & 0.763 & 87.0 \\
AlphaFold3 
  & 0.810 (–0.6) & 93.9 (+1.6pp) 
  & 0.839 (+1.0) & 94.5 (+1.6pp) 
  & 0.788 (+3.2) & 91.1 (+4.0pp) \\
\rowcolor{gray!15}\methodname (Ours) 
  & \textbf{0.824 (+1.2)} & \textbf{94.9 (+2.6pp)} 
  & \textbf{0.850 (+2.3)} & \textbf{95.7 (+2.9pp)} 
  & \textbf{0.800 (+4.8)} & \textbf{92.2 (+5.2pp)} \\
\bottomrule
\end{tabular}
}
\end{table}

In this section, we demonstrate that \methodname retains strong capability for one-step prediction.

\paragraph{Baselines} We compare these AlphaFold3 variants: (i) \textbf{AlphaFold3} \citep{abramson2024accurate} – The original configuration of AlphaFold3 employs the full set of recycling cycles and diffusion steps, serving as a strong baseline as well as the reference target that \methodname aims to approximate. (ii) \textbf{AF3 ODE} – AlphaFold3 configured with a single sampling step and a single recycling cycle, serving as a reference baseline without retraining. (iii) \textbf{AF3 TGM} – a partially distilled AlphaFold3 variant, which builds upon AF3 ODE by applying only our TGM diffusion consistency distillation without pairformer distillation. This isolates the contribution of TGM to performance under one-step sampling. (iv) \textbf{\methodname} – our fully distilled model after applying dual consistency training, which uses only 1 recycle and 1 diffusion denoising step. Both the baseline and the initialization of \methodname are derived from Protenix, an open-source reimplementation of AlphaFold3. (v) \textbf{Protenix-Mini} – We also include a lightweight variant of Protenix, which reduces the parameter count from 368M to 135M and uses 2-step ODE sampling to lower computational cost.

\paragraph{Data} For training, we use PDB entries released after September 30, 2021, organized following the Protenix scheme with identical filtering protocols. Evaluation is performed on two benchmarks: (i) \textbf{PoseBusters V2} \citep{buttenschoen2024posebusters}, a curated benchmark of recent high-quality protein–ligand crystal complexes with drug-like molecules, restricted to post-2021 releases; and (ii) the \textbf{Low Homology Recent PDB dataset} \citep{jumper2021highly, bytedance2025protenix}, containing numerous protein and nucleic acid interfaces. Introduced in AlphaFold3, we employ the Protenix open-source implementation. All entries predating the training cutoff are excluded from evaluation.

\paragraph{Metrics} On Posebusters V2, we evaluate predictions using the RMSD between predicted and experimental ligand coordinates. For each complex, we report the proportions of generated poses whose best and worst RMSDs (with respect to the ground-truth structure) fall below the thresholds of 1, 2, 3, and 5 Å. Ground truth is not used for any filtering, so this does not introduce data leakage. These metrics quantify how Dual Consistency reshapes AlphaFold3’s output distribution. On RecentPDB, we measure backbone accuracy using the TM-score \citep{biasini2013openstructure}, where values above 0.5 indicate correct folds; the success rate is defined as the proportion of structures with RMSD $<$ 2 Å; and local accuracy is assessed using lDDT \citep{mariani2013lddt}, which ranges from 0–100 and reflects residue-level geometric precision.

\begin{figure}
    \centering
    \includegraphics[width=1\linewidth]{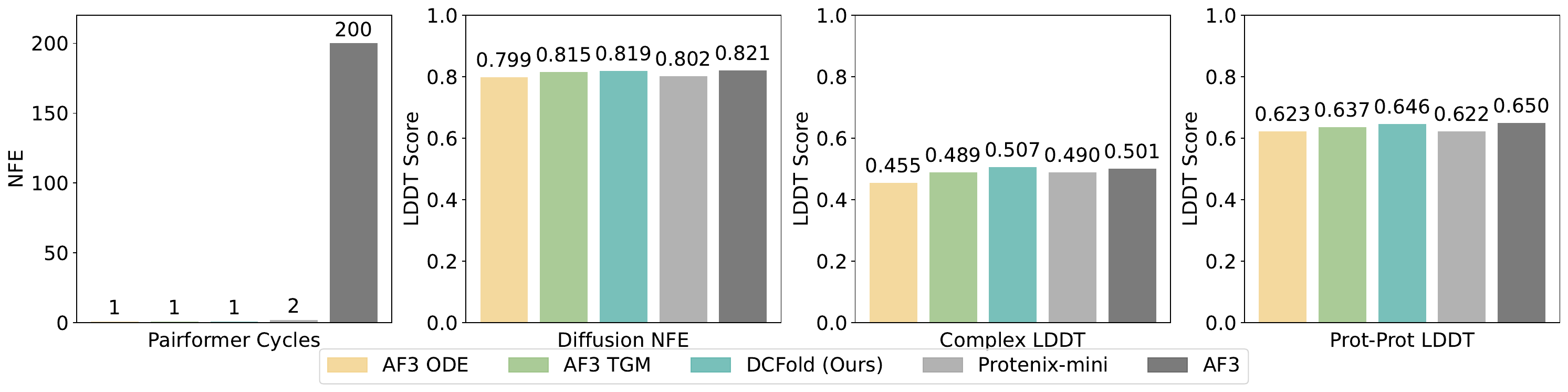}
    \caption{lDDT performance on the Recent PDB dataset.}
    \label{fig:lddt}
\end{figure}

Overall, \methodname achieves accuracy comparable to AlphaFold3 while using only a single recycle and diffusion step, demonstrating both efficiency and robustness. The results in Table~\ref{tab:posebusters-rmsd}, Table \ref{tab:tm_sr_comparison} and Figure \ref{fig:lddt} highlight these key observations:

\textbf{AlphaFold3 admits single-step generation.} With a proper choice of ODE parameters, the AF3 ODE solver is capable of generating approximately correct protein structures.

\textbf{\methodname enhances generative performance.} Training with Dual Consistency substantially improves the performance of the AF3 ODE model: across several RMSD thresholds, \methodname approaches or even matches AlphaFold3, demonstrating that the distilled model effectively recovers accuracy despite relying on only a single recycle and diffusion step.

\textbf{\methodname reshapes the distribution of generated structures.} Dual Consistency reshapes the output distribution of AlphaFold3 by effectively tightening it. This effect is reflected in the improved \textit{worst}-case RMSD, indicating more stable and reliable predictions, while the \textit{best}-case RMSD remains largely unchanged. Such a redistribution reduces extreme errors and enhances the consistency of single-step predictions, which is particularly valuable for accelerating downstream scientific workflows where both efficiency and reliability are critical.

The improvement is especially evident in Success Rate, where \methodname achieves substantially larger gains than in average TM-score. This observation further supports our claim that \methodname reshapes the distribution of generated structures. In particular, \methodname demonstrates a stronger ability than AlphaFold3 to avoid generating implausible biological complexes.

\textbf{Both components of Dual Consistency are beneficial.} In the lDDT experiments shown in Figure~\ref{fig:lddt}, \methodname delivers accuracy on par with AlphaFold3. We further conduct ablation studies disentangling the effects of Diffusion Consistency and Pairformer Consistency, and find that both components contribute complementary gains. Together, these results highlight that Dual Consistency is the key driver behind the observed improvements.

\subsection{Diversity and Confidence}
To more comprehensively characterize the performance of \methodname, we conducted an extended analysis of its structural diversity and predictive confidence on the Posebusters V2 benchmark.

\paragraph{Metrics.}
For each test sequence, we sampled five structures and computed all pairwise TM-scores among these predictions. We report the dataset-level average of these pairwise values as the \emph{Diversity} metric (lower is better). We further compute the mean pLDDT across all sampled structures as the \emph{Confidence} metric (higher is better).

\textbf{\methodname maintains strong sample diversity and confidence.}
As shown in Table~\ref{tab:diversity_confidence}, after Dual Consistency training, \methodname exhibits no substantial deviation from AlphaFold3 in either metric. Diversity shows a slight decrease, whereas confidence displays a slight increase. These trends suggest that enforcing Dual Consistency mildly concentrates the structural distribution while preserving high prediction quality.

To assess the robustness of these observations under increased sampling, we additionally evaluated: (i) 15 samples drawn under a fixed seed, and (ii) 5 random seeds with 1 sample each. Under both settings, neither AlphaFold3 nor \methodname exhibited meaningful improvements in diversity. This behavior aligns with the well-known strong conditionality of AlphaFold-series models, which tends to limit diversity gains from additional sampling alone.

Importantly, \methodname remains compatible with a broad set of diversity-enhancing strategies proposed in prior work, including sampling MSAs, clustering or masking MSA columns, and tuning dropout rates~\citep{wayment2024predicting,wallner2023afsample,kalakoti2025afsample2}. Our acceleration approach is orthogonal to these methods, and all such techniques can be directly applied to \methodnamecomma\ with expected diversity improvements comparable to those previously reported for AlphaFold3.

\begin{table}[t]
\centering
\caption{Diversity and confidence metrics on the Posebusters V2 benchmark.}
\label{tab:diversity_confidence}
\begin{tabular}{lcc}
\toprule
\textbf{Method} & \textbf{Diversity (↓)} & \textbf{Confidence (↑)} \\
\midrule
AF3 (5 samples)  & $0.9646 \pm 0.0410$ & $93.97 \pm 2.92$ \\
AF3 (15 samples) & $0.9642 \pm 0.0415$ & $93.95 \pm 2.93$ \\
AF3 (5 seeds $\times$ 1 sample) & $0.9697 \pm 0.0421$ & $93.90 \pm 3.01$ \\
\midrule
\methodname (5 samples) & $0.9701 \pm 0.0565$ & $94.14 \pm 2.97$ \\
\methodname (15 samples) & $0.9708 \pm 0.0567$ & $94.13 \pm 2.96$ \\
\methodname (5 seeds $\times$ 1 sample) & $0.9712 \pm 0.0570$ & $94.15 \pm 2.97$ \\
\bottomrule
\end{tabular}
\end{table}

\subsection{Binder Hallucination}
\label{sec:binder_exp}
After maintaining the consistency between Pairformer and Diffusion, \methodname achieves efficient inference and stable gradient backpropagation with modest computational cost. We focus on the binder hallucination task, which serves as a representative benchmark due to its stringent requirements: it demands a fully differentiable folding model, while the filtering stage eliminates a large fraction of implausible candidates. As a result, success in this setting critically depends on achieving efficient inference. Following the same hallucination strategy and filtering pipeline as BindCraft (details provided in Appendix \ref{app:binder}) \citep{pacesa2024bindcraft}, we leverage confidence scores and additional loss terms from \methodname as feedback signals for sequence evaluation. To ensure a fair comparison, folding constraints are consistently computed using the outputs of AlphaFold2, thereby avoiding potential numerical discrepancies in confidence calibration between \methodname and AlphaFold2.

\paragraph{Data} We adopt the six representative entries from \cite{cao2022design} as the design targets, namely IL2-R$\alpha$, TrkA, H3, VirB8, ALK, and LTK. They span multiple functional categories, including receptors, enzymes, transcription factors, and bacterial proteins. They have been widely adopted in prior studies as common benchmarks for design and docking tasks. For each case, we restrict binder length to 55–65 residues and perform a continuous 48-hour hallucination run. 

\paragraph{Metrics} 

We compute the Success Rate using the same two filters as BindCraft. The model-based constraint is derived from AlphaFold2’s confidence score, whereas the physics-based constraint relies on physical metrics obtained from Rosetta. Additional details are provided in Appendix \ref{app:binder}.

\begin{table}[t]
\centering
\caption{In silico success rates across six targets for binder design (values shown as physics-based constraints / model-based constraints).}
\label{tab:binder_hallucination}
\begin{tabular}{lccccccc}
\toprule
 & \textbf{IL-2R$\alpha$} & \textbf{TrkA} & \textbf{H3} & \textbf{VirB8} & \textbf{ALK} & \textbf{LTK} & \textbf{Average} \\
\midrule
BindCraft & \textbf{.38/.84} & .29/\textbf{.88} & .16/.52 & .15/.72 & \textbf{.14}/.48 & .43/.70 & .26/.69 \\
\methodname (Ours) & .37/.79 & \textbf{.31}/.84 & \textbf{.23/.71} & \textbf{.21/.85} & .12/\textbf{.54} & \textbf{.47/.93} & \textbf{.29/.78} \\
\bottomrule
\end{tabular}
\end{table}

\methodname achieves higher in silico success rates than the AF2-based BindCraft baseline across the majority of targets. With the incorporation of \methodnamecomma, AlphaFold3 can readily support binder hallucination strategies that were previously only feasible within the AlphaFold2 framework. Notably, \methodname achieves much higher success rates on several targets (e.g., H3, VirB8, and LTK), indicating that our reshaping of AlphaFold3’s output distribution translates into tangible improvements in downstream design tasks. These findings highlight that \methodname bridges the methodological gap between AlphaFold2- and AlphaFold3-based pipelines, and unlocks additional performance gains. We have added more details about the experimental results in Appendix \ref{app:binder_hallucination}. Figure \ref{fig:binder_cases} visualizes representative binder–target complexes, illustrating the interactions between the generated binders and their targets.

\subsection{Empirical Validation of TGM}
\label{sec:TGM_exp}
\begin{wraptable}{r}{0.5\textwidth}
\centering
\caption{Success Rates of Different Consistency Models on Posebusters V2.}
\label{tab:TGM_ablation}
\begin{tabular}{lcc}
\toprule
\textbf{Method} & \textbf{Time (s/step)} & \textbf{Success rate (\%)} \\
\midrule
CD  & 18.5 & 25.6\textcolor{red}{$\downarrow$} \\
sCM & 38.1 & - \\
ECM & \textbf{11.6} & 75.7\textcolor{green}{$\uparrow$} \\
\rowcolor{gray!15} TGM & \textbf{11.6} & \textbf{77.5}\textcolor{green}{$\uparrow$} \\
\bottomrule
\end{tabular}
\end{wraptable}

We conduct experiments on feasible generic consistency-model baselines, including CD \citep{song2023consistency}, sCM \citep{lu2024simplifying}, ECM \citep{geng2024consistency}, and TGM. Results on Posebusters V2 are summarized in Table~\ref{tab:TGM_ablation}.\footnote{Due to the substantial computational overhead of sCM, processing long sequences often results in out-of-memory (OOM) errors, preventing it from participating in a fair comparison.} We observe that among all runnable baselines, a naive implementation of CD leads to training collapse and severely degrades performance. Only ECM and TGM are able to enhance the performance of the diffusion module, with TGM yielding the largest performance gains. Therefore, in the following experiments, we take ECM as the representative of prior general consistency models and investigate how TGM exhibits distinct behavior on protein folding tasks. Detailed hyperparameter settings for each method are provided in Appendix \ref{app:tgm_baselines}.

\begin{wrapfigure}{r}{0.35\textwidth}
    \centering
    \includegraphics[width=0.35\textwidth]{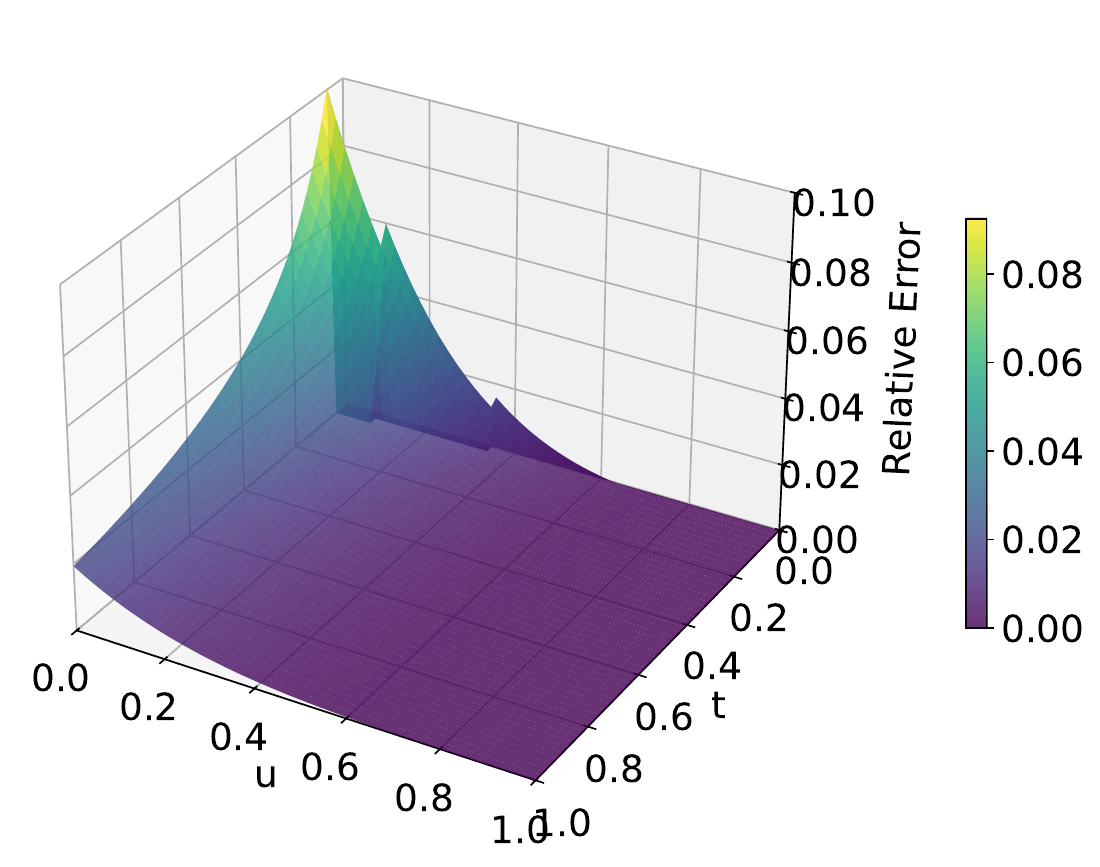}
    \caption{The relative error of the Euler solver for $r(t,u)$.}
    \vspace{-18pt}
    \label{fig:error_of_r}
\end{wrapfigure}

\begin{figure}[t]
    \centering
    \includegraphics[width=1\textwidth]{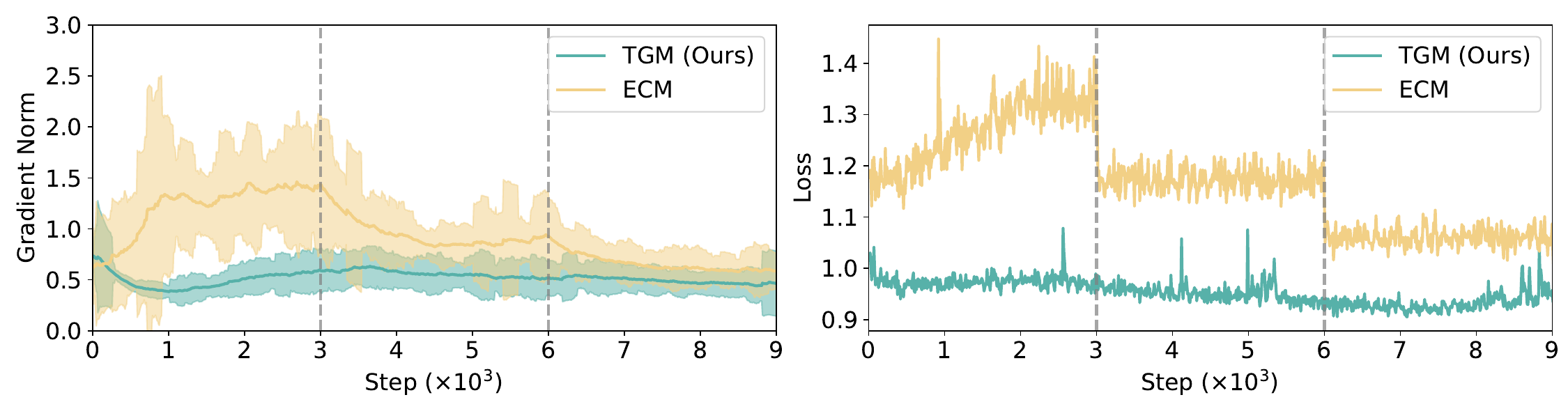}
    \caption{Gradient norm and loss curve during training for ECM and TGM.}
    \label{fig:grad_norm}
\end{figure}

We conduct an in-depth analysis of the sources of improvement introduced by TGM and present the gradient norm and loss curve throughout training in Figure \ref{fig:grad_norm}. We observe that the training dynamics of ECM exhibits poor smoothness, characterized by distinct staircase-like patterns, and is accompanied by a large gradient variance. This corroborates our hypothesis in Section \ref{sec:TGM} that classical consistency algorithms degrade under variable-length sequences. In contrast, TGM consistently maintains balanced gradients, indicating that the learning difficulty of the network remains at a fixed distance from its current capacity, effectively counteracting the adverse effects introduced by variable-length sequences.

In addition, we further assess whether the Euler method employed in TGM introduces excessive numerical error in Figure \ref{fig:error_of_r}. We observe that the error is relatively large during the early stages of training but decreases as training progresses, leading to more accurate estimates in later stages. Moreover, the error remains consistently low throughout the entire training process, indicating that our approximation is sufficiently reliable. This also explains why employing higher-order algorithms does not yield substantially greater benefits.

\section{Related Work}
\paragraph{Protein Structure Prediction} Protein structure prediction has rapidly advanced with deep learning. Classical methods such as Rosetta \citep{rohl2004protein} and co-evolutionary analysis \citep{marks2011protein, ovchinnikov2017protein} provided key insights but were limited in accuracy and scalability. The advent of deep neural networks enabled models like RaptorX \citep{xu2019distance} and trRosetta \citep{yang2020improved} to exploit large multiple sequence alignments (MSAs), setting the stage for a decisive breakthrough. AlphaFold2 \citep{jumper2021highly} combined evolutionary information with a novel attention architecture, achieving near-experimental resolution.

Efforts to reduce reliance on MSAs led to models such as ESMFold \citep{lin2022language}, OmegaFold \citep{wu2022high}, and HelixFold-Single \citep{fang2022helixfold}, which leverage protein language models for fast single-sequence prediction, albeit at lower accuracy. Extensions like AlphaFold-Multimer \citep{evans2021protein} generalized AF2 to protein–protein interactions, establishing it as a foundation model. Building on this, AlphaFold3 \citep{abramson2024accurate} introduced a diffusion-based structure module and unified biomolecular representation, enabling prediction of protein–ligand, nucleic acid, and heterogeneous complexes. Despite setting new standards in accuracy and scope, AF3’s computational overhead remains a key barrier, driving research into acceleration, distillation, and approximation \citep{cheng2022fastfold}.

\paragraph{Diffusion Acceleration} Recent advances in diffusion acceleration fall into three categories: training-free solvers, training-based distillation, and flow-based reformulations. Training-free solvers leverage higher-order integration, predictor–corrector schemes, and adaptive noise schedules to achieve high-quality generation in a few dozen steps, though performance often degrades in the extreme few-step regime \citep{song2020denoising, lu2022dpm, zhao2023unipc}. Training-based distillation compresses long diffusion chains into compact generators: progressive distillation iteratively reduces step counts, adversarial variants integrate GAN-style objectives, and Consistency Models (CMs) enforce self-consistency across time to enable single- or few-step generation with strong fidelity \citep{salimans2022progressive, sauer2024adversarial, song2023consistency}. In parallel, flow-based methods reformulate diffusion as velocity fields with straightened trajectories, allowing efficient integration with simple solvers \citep{liu2022flow, lipman2022flow}.

\section{Conclusion}
We present \methodnamecomma, a dual-consistency distillation framework that compresses AlphaFold3 into a high-fidelity single-step sampler. By jointly enforcing diffusion and Pairformer consistency and introducing the Temporal Geodesic Matching schedule, \methodname achieves stable training on variable-length protein sequences while reducing inference cost by up to \textbf{15×}. Experiments on structure prediction and binder design show that \methodname matches or surpasses AlphaFold3 in accuracy and substantially improves downstream usability, bridging AlphaFold2’s efficiency with AlphaFold3’s accuracy to enable scalable, differentiable protein design.

\section*{Ethics Statement}
This work focuses on methodological contributions to protein structure prediction and design. All experiments are conducted on publicly available datasets such as the Protein Data Bank (PDB) and established benchmarks, without involving human subjects, sensitive personal data, or animal studies. The proposed methods are intended solely for advancing computational biology research. Therefore, we do not identify any specific ethical concerns associated with this work.

\section*{Reproducibility Statement}
We disclose all training details in Section \ref{sec:dual_consistency} and Section~\ref{sec:TGM}, enabling full reproducibility of our experimental results. Moreover, we will release both the pretrained weights and the source code to ensure transparency and facilitate future research.

\section*{Acknowledgments}
The authors would thank the anonymous reviewers for reviewing the draft. This work is supported by the Natural Science Foundation of China (Grant No. 62376133), Beijing Nova Program (Grant No. 20240484682) and Wuxi Research Institute of Applied Technologies, Tsinghua University under Grant 20242001120.

\bibliography{iclr2026_conference}
\bibliographystyle{iclr2026_conference}

\clearpage
\appendix
\begin{figure}
    \label{fig:complex}
    \centering
    \includegraphics[width=1\linewidth]{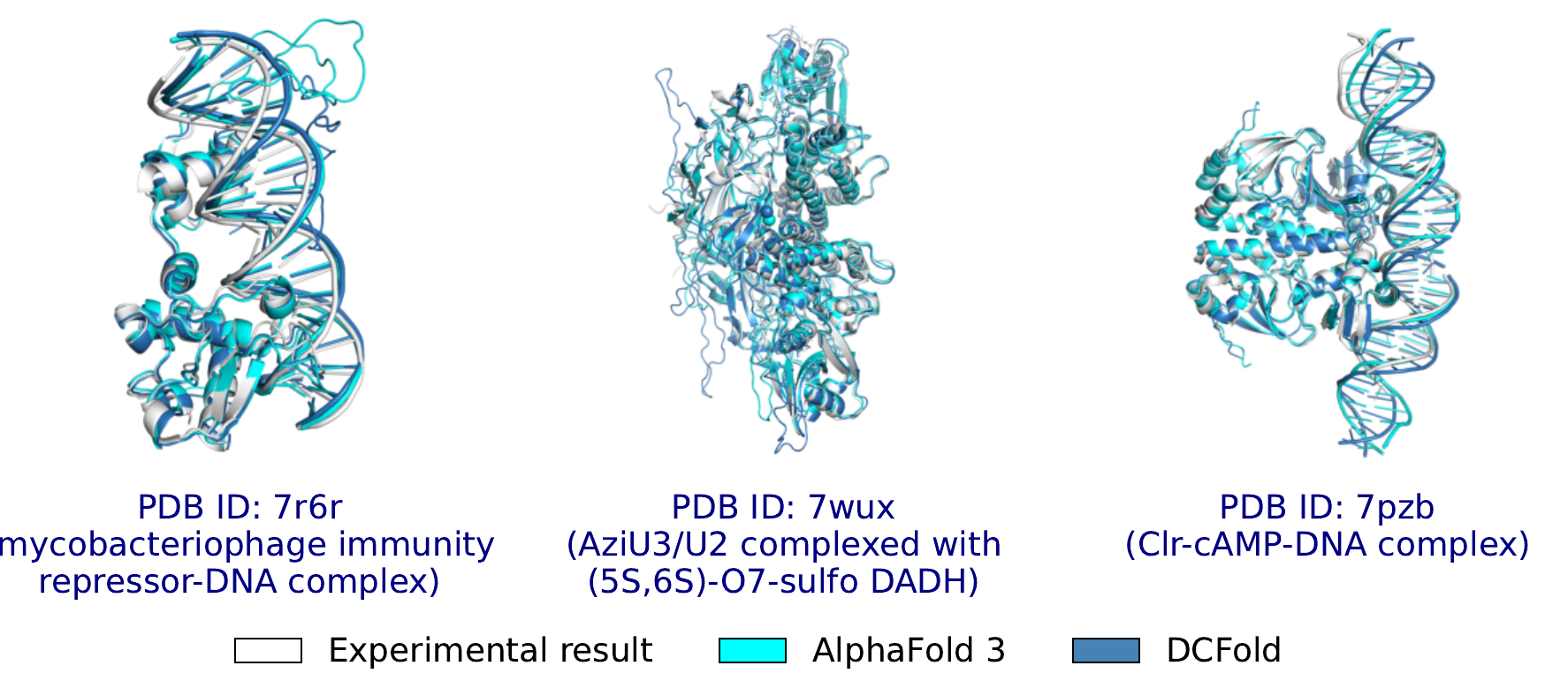}
    \caption{A structure prediction case study of \methodnamecomma, compared against AlphaFold3 and the experimental result.}
\end{figure}

\begin{figure}
    \centering
    \includegraphics[width=1\linewidth]{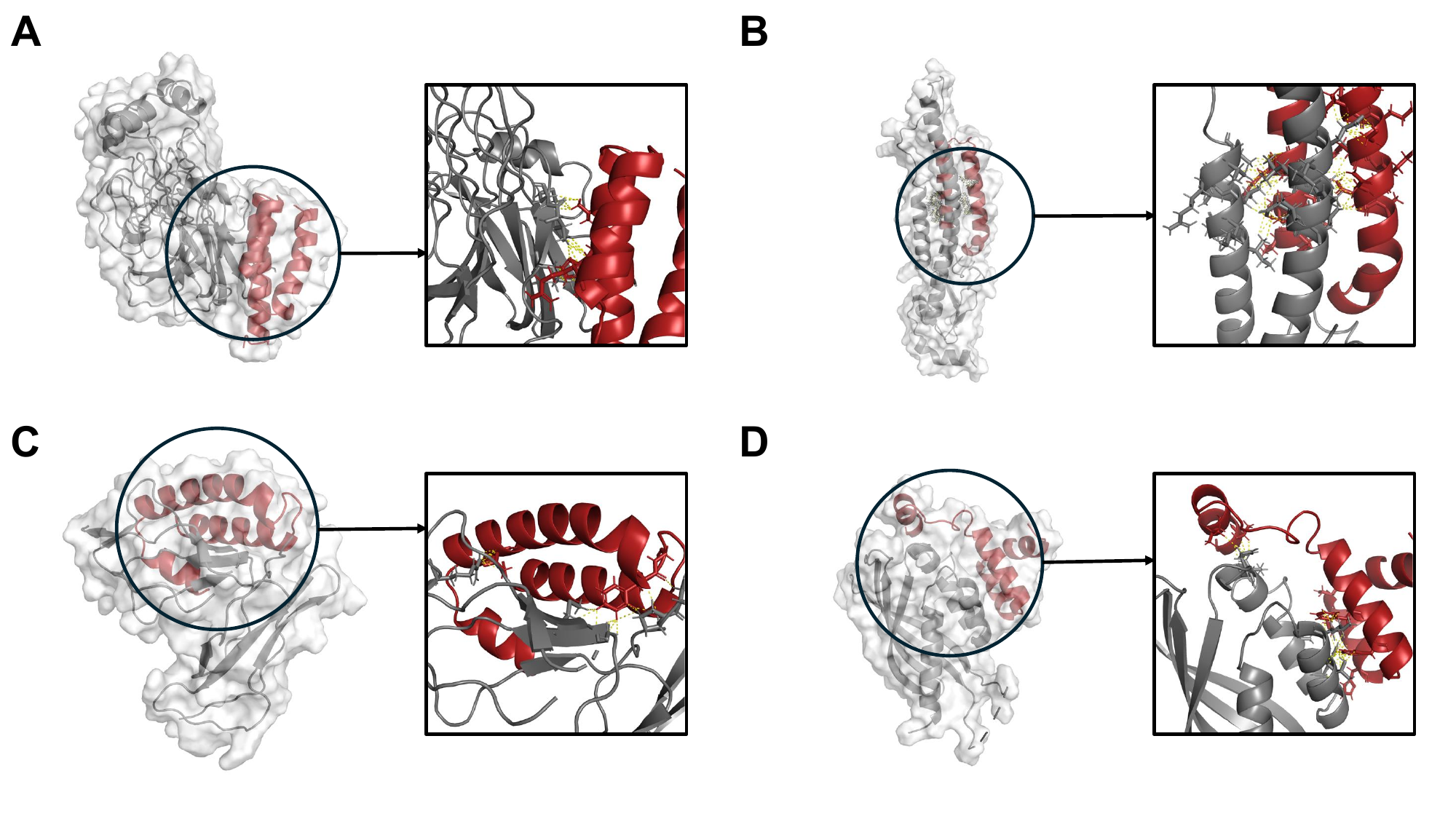}
    \caption{Examples from binder-design experiments, with targets: (A) ALK, (B) H3, (C) IL2R$\alpha$, and (D) VirB8.}
    \label{fig:binder_cases}
\end{figure}

\section{Derivation of TGM}
\subsection{Proof of Local Metric-KL Equivalence}
\label{app:metric-kl_proof}
We investigate the KL divergence between two distributions defined on the manifold $\mathcal{M}_t$:
\begin{equation}
    \KL(p_r \Vert p_t) = \int p_r(x) \log \frac{p_r(x)}{p_t(x)}\,dx = \int p_r(x)\left[\log p_r(x) - \log p_t(x)\right]\,dx
\end{equation}

We perform a Taylor expansion of $\log p_{t-\Delta t}(x)$ and substitute the result into the KL divergence.
\begin{equation}
    \log p_{t-\Delta t}(x) = \log p_t(x) - \Delta t \frac{\partial}{\partial t}\log p_t(x)+\frac{1}{2} (\Delta t)^2
    \frac{\partial^2}{\partial t^2}\log p_t(x) + \mathcal{O}\left((\Delta t)^3\right)
\end{equation}

Substituting it into the KL divergence yields:
\begin{align}
    \KL(p_r \Vert p_t) = & \int p_r(x)\left[-\Delta t\frac{\partial}{\partial t} \log p_t(x) + \frac{1}{2}(\Delta t)^2 \frac{\partial^2}{\partial t^2} \log p_t(x) + \mathcal{O}\left((\Delta t)^3\right)\right]\,dx \nonumber \\
    = & \int \left[p_t(x)-\Delta t \frac{\partial}{\partial t} p_t(x) + \frac{1}{2} (\Delta t)^2 \frac{\partial^2}{\partial t^2} p_t(x) + \mathcal{O}\left((\Delta t)^3\right)\right] \cdot \nonumber \\ & \left[-\Delta t\frac{\partial}{\partial t} \log p_t(x) + \frac{1}{2}(\Delta t)^2 \frac{\partial^2}{\partial t^2} \log p_t(x) + \mathcal{O}\left((\Delta t)^3\right)\right] \,dx
\end{align}

The first-order term vanishes:
\begin{equation}
    -\Delta t \int p_t(x) \frac{\partial}{\partial t} \log p_t(x) \,dx = 0,
\end{equation}

while the second-order term takes the following form:
\begin{equation}
    \frac{(\Delta t)^2}{2} \int p_t(x) \frac{\partial^2}{\partial t^2} \log p_t(x) \,dx + (\Delta t)^2\int \frac{\partial}{\partial t}p_t(x)\frac{\partial}{\partial t}\log p_t(x) \,dx
\end{equation}

The term on the right-hand side is given by
\begin{equation}
    (\Delta t)^2 \int p_t(x)\left[\frac{\partial}{\partial t} \log _t(x)\right]^2 \,dx = (\Delta t)^2 \mathcal{I}(t)
\end{equation}

The simplification of the left-hand side relies on the property that the integral of the score function vanishes:
\begin{equation}
    0=\frac{\partial}{\partial t} \cdot 0 = \frac{\partial}{\partial t}\int p_t(x)\frac{\partial}{\partial t}\log p_t(x) \,dx = \int \frac{\partial}{\partial t} p_t(x) \frac{\partial}{\partial t} \log p_t(x) \,dx + \int p_t(x) \frac{\partial^2}{\partial t^2} \log p_t(x) \,dx
\end{equation}

Thus, the term on the left-hand side can also be expressed in terms of $\mathcal{I}(t)$:
\begin{equation}
    \int p_t(x)\frac{\partial^2}{\partial t^2} \log p_t(x) \,dx = - \int \frac{\partial}{\partial t} p_t(x) \frac{\partial}{\partial t} \log p_t(x) \,dx = -\mathcal{I}(t)
\end{equation}

Thus, the second-order term implicitly encodes the temporal Fisher information $-\frac{(\Delta t)^2}{2} \mathcal{I}(t) + (\Delta t)^2\mathcal{I}(t) = \frac{(\Delta t)^2}{2} \mathcal{I}(t)$, that is $\KL\left(p_r(x) \Vert p_t(x)\right) = \frac{(\Delta t)^2}{2} \mathcal{I}(t) + \mathcal{O}\left((\Delta t)^3\right)$. With this, local metric-KL equivalence becomes evident.

\subsection{Temporal Fisher Information in EDM}
We assume the forward process of diffusion is defined as $p_t(x|x_0) = \mathcal{N}(x;\mu=\alpha(t) x_0, \sigma^2(t) I), \mathcal{I}(t)=\E_{p_t(x)} \left[\left(\frac{\partial}{\partial t} \log p_t(x)\right)^2\right] = \E_{x_0 \sim p_{\text{data}}} \E_{p_t(x|x_0)} \left[\left(\frac{\partial}{\partial t} \log p_t(x)\right)^2\right]$

We employ a multivariate Gaussian distribution with dimensionality $D$: $p(x) = \frac{1}{(2\pi)^{D/2} |\Sigma|^{1/2}} \exp \left(-\frac{1}{2} (x-\mu)^\top \Sigma^{-1} (x-\mu)\right)$, $\Sigma = \sigma^2(t) I, |\Sigma| = |\sigma^2(t)I| = \sigma^{2D}(t), \Sigma^{-1} = \left(\sigma^2(t)I\right)^{-1} = \sigma^{-2}(t)I$, which yields the following simplification:

\begin{equation}
    p_t(x|x_0) = \frac{1}{(2\pi)^{D/2} \sigma^D(t)} \exp \left(-\frac{\Vert x-\alpha(t)x_0 \Vert^2}{2\sigma^2(t)}\right)
\end{equation}
\begin{equation}
    \log p_t(x|x_0) = -\frac{D}{2} \log (2\pi) -D\log \sigma(t) - \frac{\Vert x-\alpha(t)x_0 \Vert^2}{2\sigma^2(t)}
\end{equation}
\begin{align}
    \frac{\partial}{\partial t} \log p_t(x|x_0) & = -D \frac{\dot{\sigma}(t)}{\sigma(t)} - \left[-\frac{\dot{\sigma}(t)}{\sigma^3(t)} \Vert x-\alpha(t)x_0\Vert^2 - \frac{1}{2\sigma^2(t)} \left(-2\dot{\alpha}(t)\left(x-\alpha(t)x_0\right)^\top x_0 \right)\right] \nonumber \\
    & = -D \frac{\dot{\sigma}(t)}{\sigma(t)} + \frac{\dot{\sigma}(t)}{\sigma^3(t)} \Vert x-\alpha(t)x_0\Vert^2 + \frac{\dot{\alpha}(t)}{\sigma^2(t)} \left(x-\alpha(t)x_0\right)^\top x_0 \nonumber \\
    & = -D \frac{\dot{\sigma}(t)}{\sigma(t)} + \frac{\dot{\sigma}(t)}{\sigma^3(t)}\sigma^2(t)\Vert z\Vert^2 + \frac{\dot{\alpha}(t)}{\sigma^2(t)} \left(\sigma(t)z\right)^\top x_0 \nonumber \\
    & = -D \frac{\dot{\sigma}(t)}{\sigma(t)} + \frac{\dot{\sigma}(t)}{\sigma(t)}\Vert z\Vert^2+\frac{\dot{\alpha}(t)}{\sigma(t)} z^\top x_0 \nonumber \\
    & = \frac{\dot{\sigma}(t)}{\sigma(t)} (\Vert z\Vert^2 - D) + \frac{\dot{\alpha}(t)}{\sigma(t)}z^\top x_0
\end{align}

Thus, $\mathcal{I}(t)$ can be decomposed into three components:
\begin{align}
    \mathcal{I}(t) & = \E_{x_0\sim p_{\text{data}}} \E_{p_t(x|x_0)} \left[\left(\frac{\dot{\sigma}(t)}{\sigma(t)} (\Vert z\Vert^2 - D) + \frac{\dot{\alpha}(t)}{\sigma(t)} z^\top x_0\right)^2\right] \nonumber \\ 
    & = \E_{x_0\sim p_{\text{data}}} \E_{z} \left[\left( \frac{\dot{\sigma}(t)}{\sigma(t)}\right)^2(\Vert z\Vert^2 - D)^2 + \left(\frac{\dot{\alpha}(t)}{\sigma(t)}\right)^2 (z^\top x_0)^2 + 2\cdot \frac{\dot{\sigma}(t)\dot{\alpha}(t)}{\sigma^2(t)}(\Vert z\Vert^2 - D)(z^\top x_0)\right]
\end{align}

Since the first term follows a chi-squared distribution $\Vert z\Vert^2 = \sum_i z_i^2 \sim \chi^2(D)$, in this part, we introduce the data dimension $D$: $\E[\Vert z\Vert^2]=D, \E\left[(\Vert z\Vert^2 - D)^2\right] = \Var[\Vert z\Vert^2] = 2D$

the second term is $\E\left[(z^\top x_0)^2\right] = \E\left[\left(\sum_i z_i (x_0)_i\right)\left(\sum_j z_j (x_0)_j\right)\right] = \sum_{i,j} (x_0)_i(x_0)_j\delta_{ij} = \Vert x_0\Vert^2$

The third term, namely the cross-term, vanishes: $\E\left[(\Vert z\Vert^2 - D)(z^\top x_0)\right] = \E\left[\Vert z\Vert^2\cdot (z^\top x_0)\right] - D\cdot \E[z^\top x_0] = \E\left[(\sum_i z_i^2)(\sum_j z_j(x_0)_j)\right] = \sum_{i,j} (x_0)_j \E[z_i^2 z_j] = 0$

\begin{align}
    \mathcal{I}(t) =
\E_{x_0\sim p_{\text{data}}}
\left[
\left(\frac{\dot{\sigma}(t)}{\sigma(t)}\right)^2 \cdot 2D
+
\left(\frac{\dot{\alpha}(t)}{\sigma(t)}\right)^2
\cdot \Vert x_0\Vert^2
\right]
\end{align}

In most prior works, due to the effect of data normalization, we can assume that $\E[x_0] = 0$, and therefore $\Vert x_0\Vert^2$ can be expressed in terms of $\Var[x_0]$.

In the EDM framework, $\alpha(t)=1, \sigma(t) = \sigma_{\text{data}}\cdot \left(s_{\text{max}}^{1/p} + (1-t)\cdot (s_{\text{min}}^{1/p} - s_{\text{max}}^{1/p})\right)^p$. This yields a more concise expression for $I(t)$:

\begin{align}
    \mathcal{I}(t)
=
\left(\frac{\dot{\sigma}(t)}{\sigma(t)}\right)^2 \cdot 2D
=
\frac{
  2D\cdot p^2
  \cdot
  \left(s_{\text{max}}^{1/p}-s_{\text{min}}^{1/p}\right)^2
}{
  \left[
    s_{\text{max}}^{1/p}
    +(1-t)
    \left(s_{\text{min}}^{1/p}-s_{\text{max}}^{1/p}\right)
  \right]^2
}
\end{align}

\section{Implementation Details}
\subsection{Training Configuration}
To ensure clarity and reproducibility, we provide a detailed description of the training setup. Our full training pipeline was executed on a cluster equipped with 64 NVIDIA H800 GPUs, corresponding to an effective batch size of 64. Stage~1 focuses on learning diffusion consistency. \methodname was trained for approximately 40 hours, spanning a total of 9{,}000 optimization steps. This stage establishes the foundational generative capabilities leveraged in subsequent training. Stage~2 aims to refine the structural reasoning components through Pairformer consistency training. This phase required around 7 hours of computation and was conducted for 1{,}500 steps. The shorter duration reflects both the stability provided by Stage~1 and the efficiency of fine-tuning the Pairformer module.

\subsection{Binder Hallucination}
\label{app:binder}
After initial binder design with \methodnamecomma, sequences are refined to improve stability and solubility using ProteinMPNN with soluble weights, while preserving residues within 4 Å of the target interface. For each binder, 20 variants are generated at temperature 0.1 with no backbone noise. These sequences are re-predicted using the AF2 monomer model (3 recycles, 2 template-based models) in single-sequence mode to validate structural robustness. Resulting complexes are energy-minimized with Rosetta FastRelax (200 iterations) and evaluated using InterfaceAnalyzer with sidechain and backbone movement. Final designs are filtered using predefined thresholds (pLDDT $> 0.8$, i\_pTM $> 0.5$, i\_pAE $< 0.35$, shape complementarity $> 0.55$, $< 3$ unsaturated H-bonds, binder surface hydrophobicity $< 35\%$, RMSD $< 3.5$ Å), yielding a high-confidence set of candidates.

We evaluate binder quality using two constraint sets. Model-based Constraints are derived from AlphaFold2 confidence outputs, requiring pLDDT $> 0.8$, interface pTM $> 0.5$, global pTM $> 0.45$, and interface pAE $< 0.4$. Physics-based Constraints are based on physical interface metrics from Rosetta, including shape complementarity $> 0.5$, dSASA $> 1$, $>6$ interface residues, $>2$ interface hydrogen bonds, surface hydrophobicity $< 0.37$, and $<6$ unsaturated hydrogen bonds. All metrics are aligned with the filters used in BindCraft.

\subsection{Hyperparameter Settings for Consistency Model Baselines}
\label{app:tgm_baselines}
For completeness, we provide the implementation details of all baselines considered in our experiments:
\begin{itemize}
    \item CD: Mean squared error (MSE) as the metric function with a weight decay rate of $\eta=0.995$.
    \item sCM: $H=2000$ warm-up iterations.
    \item ECM: $q=2.0$, $b=0.1$, $d=3000$, and $k=4.0$.
    \item TGM: Hyperparameter search yields $C_0 = 32$ and $\beta = 2$. In addition, we inherit the exponential decay scheduling parameters from AlphaFold3’s EDM configuration, with $p = 7$, $s_{\max} = 160$, and $s_{\min} = 4 \times 10^{-4}$.
\end{itemize}
For all methods, we set the weighting function to $1$.

\section{Experiment Details}
\subsection{Runtime Characteristics Across Sequence Lengths}
To comprehensively assess the efficiency of \methodnamecomma, we report detailed bin-wise runtime statistics on the Posebusters V2 benchmark. Since AlphaFold3 supports folding protein-ligand complexes, we use the total number of input tokens for each test entry as the length metric and partition sequences into bins of size 128. The average inference time for each bin is summarized in Table~\ref{tab:runtime-bins}.

\begin{table}[t]
\centering
\caption{Average inference time of AlphaFold3 and DCFold across token bins.}
\label{tab:runtime-bins}
\begin{tabular}{lcc}
\toprule
\textbf{\#Tokens} & \textbf{AlphaFold3 Avg Time (s)} & \textbf{\methodname Avg Time (s)} \\
\midrule
$\leq$ 255        & 92.63  & 3.76  \\
256--383      & 103.31 & 5.77  \\
384--511      & 112.35 & 7.17  \\
512--639      & 126.41 & 10.87 \\
640--767      & 142.78 & 14.65 \\
768--895      & 169.20 & 20.02 \\
$\geq$ 896       & 212.12 & 27.40 \\
\bottomrule
\end{tabular}
\end{table}

Both AlphaFold3 and \methodname exhibit increasing runtime as sequence length grows. However, the relative acceleration provided by \methodname is most pronounced for short sequences, where it achieves up to a \textbf{24$\times$} speedup. For moderately long sequences, \methodname still provides more than \textbf{7.7$\times$} acceleration, demonstrating consistent efficiency gains across all token ranges.

We hypothesize that this trend stems from the differing computational bottlenecks of the two methods. The reduction in Diffusion NFE afforded by \methodname yields a significantly larger improvement factor compared to the reduction in Pairformer cycles. As sequence length increases, the Pairformer component becomes the dominant cost, diminishing the relative impact of the diffusion speedup. Conversely, in shorter sequences, the Pairformer bottleneck is less pronounced, enabling the diffusion efficiency gains to translate directly into substantial end-to-end acceleration.

\subsection{Binder Hallucination}
\label{app:binder_hallucination}
We conducted experiments on a single H800 GPU. On the targets used in Table \ref{tab:binder_hallucination}, the average GPU time for one full hallucination with BindCraft is 138s, while \methodname requires 105s. Since we follow the same pipeline as BindCraft, the total serial runtime also includes the time for ProteinMPNN and the re-prediction step in addition to the design model’s GPU time. We also provide the total number of designs generated in our experiments in Table \ref{tab:num_binder_samples}. Overall, \methodname attains slightly better efficiency while producing a comparable number of samples, ensuring a fair comparison.

\begin{table}[t]
\centering
\caption{The total number of generated samples in the binder hallucination experiments.}
\label{tab:num_binder_samples}
\begin{tabular}{lccccccc}
\toprule
& \textbf{IL-2R$\alpha$} & \textbf{TrkA} & \textbf{H3} & \textbf{VirB8} & \textbf{ALK} & \textbf{LTK} \\
\hline
\addlinespace[2pt]
BindCraft & 312 & 243 & 269 & 347 & 188 & 348 \\
\methodname (Ours) & 375 & 256 & 295 & 439 & 177 & 402 \\
\bottomrule
\end{tabular}
\end{table}

Our binder design benchmark features six protein targets. Table~\ref{tab:binder_detail} shows the details of the targets.

\begin{table}[t]
\centering
\caption{Detailed information of binder targets in the binder hallucination experiments.}
\label{tab:miniprotein_details}
\label{tab:binder_detail}
\begin{tabular}{ccp{3.5cm}p{6cm}}
\toprule
\textbf{Target} & \textbf{PDB ID} & \textbf{Family} & \textbf{Description} \\
\midrule
ALK & 7NWZ & Immune receptor & Neural receptor tyrosine kinase involved in development \\
H3 & 3ZTJ & Receptor tyrosine kinase & Core nucleosomal histone in eukaryotic chromatin \\
IL2R$\alpha$ & 1Z92 & Histone protein & Component of the interleukin-2 receptor complex in the immune system\\
LTK & 7NX0 & Bacterial secretion system protein & Homolog of ALK expressed in various tissues \\
TrkA & 2IFG & Receptor tyrosine kinase & Neurotrophic signaling receptor activated by NGF \\
VirB8 & 4O3V & Receptor tyrosine kinase & Structural protein of the type IV secretion system in Gram-negative bacteria \\
\bottomrule
\end{tabular}
\end{table}

\section*{The Use of Large Language Models (LLMs)}
We use large language models (LLMs) solely for auxiliary editing purposes, including spelling correction and minor grammatical adjustments. Importantly, LLMs are not involved in the conception of research ideas or the development of code. We disclose this usage explicitly to ensure transparency in our work.

\end{document}

%% file: math_commands.tex
\usepackage{amsmath,amsfonts,bm}

\def\eqref#1{equation~\ref{#1}}
\def\1{\bm{1}}

\DeclareMathAlphabet{\mathsfit}{\encodingdefault}{\sfdefault}{m}{sl}
\SetMathAlphabet{\mathsfit}{bold}{\encodingdefault}{\sfdefault}{bx}{n}

\newcommand{\E}{\mathbb{E}}

\newcommand{\KL}{D_{\mathrm{KL}}}
\newcommand{\Var}{\mathrm{Var}}

%% file: iclr2026_conference.bib
@article{abramson2024accurate,
  title={Accurate structure prediction of biomolecular interactions with AlphaFold 3},
  author={Abramson, Josh and Adler, Jonas and Dunger, Jack and Evans, Richard and Green, Tim and Pritzel, Alexander and Ronneberger, Olaf and Willmore, Lindsay and Ballard, Andrew J and Bambrick, Joshua and others},
  journal={Nature},
  volume={630},
  number={8016},
  pages={493--500},
  year={2024},
  publisher={Nature Publishing Group UK London}
}

@article{alhumaid2024reliability,
  title={Reliability of AlphaFold2 models in virtual drug screening: a focus on selected class A GPCRs},
  author={Alhumaid, Nada K and Tawfik, Essam A},
  journal={International Journal of Molecular Sciences},
  volume={25},
  number={18},
  pages={10139},
  year={2024},
  publisher={MDPI}
}

@article{baselious2024comparative,
  title={Comparative structure-based virtual screening utilizing optimized AlphaFold model identifies selective HDAC11 inhibitor},
  author={Baselious, Fady and Hilscher, Sebastian and Robaa, Dina and Barinka, Cyril and Schutkowski, Mike and Sippl, Wolfgang},
  journal={International Journal of Molecular Sciences},
  volume={25},
  number={2},
  pages={1358},
  year={2024},
  publisher={MDPI}
}

@article{bennett2023improving,
  title={Improving de novo protein binder design with deep learning},
  author={Bennett, Nathaniel R and Coventry, Brian and Goreshnik, Inna and Huang, Buwei and Allen, Aza and Vafeados, Dionne and Peng, Ying Po and Dauparas, Justas and Baek, Minkyung and Stewart, Lance and others},
  journal={Nature Communications},
  volume={14},
  number={1},
  pages={2625},
  year={2023},
  publisher={Nature Publishing Group UK London}
}

@article{biasini2013openstructure,
  title={OpenStructure: an integrated software framework for computational structural biology},
  author={Biasini, Marco and Schmidt, Tobias and Bienert, Stefan and Mariani, Valerio and Studer, Gabriel and Haas, J{\"u}rgen and Johner, Niklaus and Schenk, Andreas Daniel and Philippsen, Ansgar and Schwede, Torsten},
  journal={Biological crystallography},
  volume={69},
  number={5},
  pages={701--709},
  year={2013},
  publisher={International Union of Crystallography}
}

@article{buttenschoen2024posebusters,
  title={PoseBusters: AI-based docking methods fail to generate physically valid poses or generalise to novel sequences},
  author={Buttenschoen, Martin and Morris, Garrett M and Deane, Charlotte M},
  journal={Chemical Science},
  volume={15},
  number={9},
  pages={3130--3139},
  year={2024},
  publisher={Royal Society of Chemistry}
}

@article{bytedance2025protenix,
  title={Protenix - Advancing Structure Prediction Through a Comprehensive AlphaFold3 Reproduction},
  author={ByteDance AML AI4Science Team and Chen, Xinshi and Zhang, Yuxuan and Lu, Chan and Ma, Wenzhi and Guan, Jiaqi and Gong, Chengyue and Yang, Jincai and Zhang, Hanyu and Zhang, Ke and Wu, Shenghao and Zhou, Kuangqi and Yang, Yanping and Liu, Zhenyu and Wang, Lan and Shi, Bo and Shi, Shaochen and Xiao, Wenzhi},
  year={2025},
  journal={bioRxiv},
  publisher={Cold Spring Harbor Laboratory},
  doi={10.1101/2025.01.08.631967},
  URL={https://www.biorxiv.org/content/early/2025/01/11/2025.01.08.631967},
  elocation-id={2025.01.08.631967},
  eprint={https://www.biorxiv.org/content/early/2025/01/11/2025.01.08.631967.full.pdf},
}

@article{cao2022design,
  title={Design of protein-binding proteins from the target structure alone},
  author={Cao, Longxing and Coventry, Brian and Goreshnik, Inna and Huang, Buwei and Sheffler, William and Park, Joon Sung and Jude, Kevin M and Markovi{\'c}, Iva and Kadam, Rameshwar U and Verschueren, Koen HG and others},
  journal={Nature},
  volume={605},
  number={7910},
  pages={551--560},
  year={2022},
  publisher={Nature Publishing Group UK London}
}

@article{cheng2022fastfold,
  title={Fastfold: Reducing alphafold training time from 11 days to 67 hours},
  author={Cheng, Shenggan and Zhao, Xuanlei and Lu, Guangyang and Fang, Jiarui and Yu, Zhongming and Zheng, Tian and Wu, Ruidong and Zhang, Xiwen and Peng, Jian and You, Yang},
  journal={arXiv preprint arXiv:2203.00854},
  year={2022}
}

@article{evans2021protein,
  title={Protein complex prediction with AlphaFold-Multimer},
  author={Evans, Richard and O’Neill, Michael and Pritzel, Alexander and Antropova, Natasha and Senior, Andrew and Green, Tim and {\v{Z}}{\'\i}dek, Augustin and Bates, Russ and Blackwell, Sam and Yim, Jason and others},
  journal={biorxiv},
  pages={2021--10},
  year={2021},
  publisher={Cold Spring Harbor Laboratory}
}

@article{fang2022helixfold,
  title={Helixfold-single: Msa-free protein structure prediction by using protein language model as an alternative},
  author={Fang, Xiaomin and Wang, Fan and Liu, Lihang and He, Jingzhou and Lin, Dayong and Xiang, Yingfei and Zhang, Xiaonan and Wu, Hua and Li, Hui and Song, Le},
  journal={arXiv preprint arXiv:2207.13921},
  year={2022}
}

@article{frank2024scalable,
  title={Scalable protein design using optimization in a relaxed sequence space},
  author={Frank, Christopher and Khoshouei, Ali and Fu$\beta$, Lara and Schiwietz, Dominik and Putz, Dominik and Weber, Lara and Zhao, Zhixuan and Hattori, Motoyuki and Feng, Shihao and de Stigter, Yosta and others},
  journal={Science},
  volume={386},
  number={6720},
  pages={439--445},
  year={2024},
  publisher={American Association for the Advancement of Science}
}

@article{geng2024consistency,
  title={Consistency models made easy},
  author={Geng, Zhengyang and Pokle, Ashwini and Luo, William and Lin, Justin and Kolter, J Zico},
  journal={arXiv preprint arXiv:2406.14548},
  year={2024}
}

@article{ho2020denoising,
  title={Denoising diffusion probabilistic models},
  author={Ho, Jonathan and Jain, Ajay and Abbeel, Pieter},
  journal={Advances in neural information processing systems},
  volume={33},
  pages={6840--6851},
  year={2020}
}

@article{jendrusch2025alphadesign,
  title={AlphaDesign: A de novo protein design framework based on AlphaFold},
  author={Jendrusch, Michael A and Yang, Alessio LJ and Cacace, Elisabetta and Bobonis, Jacob and Voogdt, Carlos GP and Kaspar, Sarah and Schweimer, Kristian and Perez-Borrajero, Cecilia and Lapouge, Karine and Scheurich, Jacob and others},
  journal={Molecular Systems Biology},
  pages={1--24},
  year={2025},
  publisher={Nature Publishing Group UK London}
}

@article{jumper2021highly,
  title={Highly accurate protein structure prediction with AlphaFold},
  author={Jumper, John and Evans, Richard and Pritzel, Alexander and Green, Tim and Figurnov, Michael and Ronneberger, Olaf and Tunyasuvunakool, Kathryn and Bates, Russ and {\v{Z}}{\'\i}dek, Augustin and Potapenko, Anna and others},
  journal={nature},
  volume={596},
  number={7873},
  pages={583--589},
  year={2021},
  publisher={Nature Publishing Group UK London}
}

@article{kalakoti2025afsample2,
  title={AFsample2 predicts multiple conformations and ensembles with AlphaFold2},
  author={Kalakoti, Yogesh and Wallner, Bj{\"o}rn},
  journal={Communications Biology},
  volume={8},
  number={1},
  pages={373},
  year={2025},
  publisher={Nature Publishing Group UK London}
}

@article{karras2022elucidating,
  title={Elucidating the design space of diffusion-based generative models},
  author={Karras, Tero and Aittala, Miika and Aila, Timo and Laine, Samuli},
  journal={Advances in neural information processing systems},
  volume={35},
  pages={26565--26577},
  year={2022}
}

@article{li2023machine,
  title={Machine learning optimization of candidate antibody yields highly diverse sub-nanomolar affinity antibody libraries},
  author={Li, Lin and Gupta, Esther and Spaeth, John and Shing, Leslie and Jaimes, Rafael and Engelhart, Emily and Lopez, Randolph and Caceres, Rajmonda S and Bepler, Tristan and Walsh, Matthew E},
  journal={Nature communications},
  volume={14},
  number={1},
  pages={3454},
  year={2023},
  publisher={Nature Publishing Group UK London}
}

@article{lin2022language,
  title={Language models of protein sequences at the scale of evolution enable accurate structure prediction},
  author={Lin, Zeming and Akin, Halil and Rao, Roshan and Hie, Brian and Zhu, Zhongkai and Lu, Wenting and dos Santos Costa, Allan and Fazel-Zarandi, Maryam and Sercu, Tom and Candido, Sal and others},
  journal={BioRxiv},
  volume={2022},
  pages={500902},
  year={2022}
}

@article{lipman2022flow,
  title={Flow matching for generative modeling},
  author={Lipman, Yaron and Chen, Ricky TQ and Ben-Hamu, Heli and Nickel, Maximilian and Le, Matt},
  journal={arXiv preprint arXiv:2210.02747},
  year={2022}
}

@article{liu2022flow,
  title={Flow straight and fast: Learning to generate and transfer data with rectified flow},
  author={Liu, Xingchao and Gong, Chengyue and Liu, Qiang},
  journal={arXiv preprint arXiv:2209.03003},
  year={2022}
}

@article{lu2022dpm,
  title={Dpm-solver: A fast ode solver for diffusion probabilistic model sampling in around 10 steps},
  author={Lu, Cheng and Zhou, Yuhao and Bao, Fan and Chen, Jianfei and Li, Chongxuan and Zhu, Jun},
  journal={Advances in neural information processing systems},
  volume={35},
  pages={5775--5787},
  year={2022}
}

@article{lu2024simplifying,
  title={Simplifying, stabilizing and scaling continuous-time consistency models},
  author={Lu, Cheng and Song, Yang},
  journal={arXiv preprint arXiv:2410.11081},
  year={2024}
}

@article{mariani2013lddt,
  title={lDDT: a local superposition-free score for comparing protein structures and models using distance difference tests},
  author={Mariani, Valerio and Biasini, Marco and Barbato, Alessandro and Schwede, Torsten},
  journal={Bioinformatics},
  volume={29},
  number={21},
  pages={2722--2728},
  year={2013},
  publisher={Oxford University Press}
}

@article{marks2011protein,
  title={Protein 3D structure computed from evolutionary sequence variation},
  author={Marks, Debora S and Colwell, Lucy J and Sheridan, Robert and Hopf, Thomas A and Pagnani, Andrea and Zecchina, Riccardo and Sander, Chris},
  journal={PloS one},
  volume={6},
  number={12},
  pages={e28766},
  year={2011},
  publisher={Public Library of Science San Francisco, USA}
}

@article{ovchinnikov2017protein,
  title={Protein structure determination using metagenome sequence data},
  author={Ovchinnikov, Sergey and Park, Hahnbeom and Varghese, Neha and Huang, Po-Ssu and Pavlopoulos, Georgios A and Kim, David E and Kamisetty, Hetunandan and Kyrpides, Nikos C and Baker, David},
  journal={Science},
  volume={355},
  number={6322},
  pages={294--298},
  year={2017},
  publisher={American Association for the Advancement of Science}
}

@article{pacesa2024bindcraft,
  title={BindCraft: one-shot design of functional protein binders},
  author={Pacesa, Martin and Nickel, Lennart and Schellhaas, Christian and Schmidt, Joseph and Pyatova, Ekaterina and Kissling, Lucas and Barendse, Patrick and Choudhury, Jagrity and Kapoor, Srajan and Alcaraz-Serna, Ana and others},
  journal={bioRxiv},
  pages={2024--09},
  year={2024},
  publisher={Cold Spring Harbor Laboratory}
}

@incollection{rohl2004protein,
  title={Protein structure prediction using Rosetta},
  author={Rohl, Carol A and Strauss, Charlie EM and Misura, Kira MS and Baker, David},
  booktitle={Methods in enzymology},
  volume={383},
  pages={66--93},
  year={2004},
  publisher={Elsevier}
}

@inproceedings{rombach2022high,
  title={High-resolution image synthesis with latent diffusion models},
  author={Rombach, Robin and Blattmann, Andreas and Lorenz, Dominik and Esser, Patrick and Ommer, Bj{\"o}rn},
  booktitle={Proceedings of the IEEE/CVF conference on computer vision and pattern recognition},
  pages={10684--10695},
  year={2022}
}

@article{salimans2022progressive,
  title={Progressive distillation for fast sampling of diffusion models},
  author={Salimans, Tim and Ho, Jonathan},
  journal={arXiv preprint arXiv:2202.00512},
  year={2022}
}

@inproceedings{sauer2024adversarial,
  title={Adversarial diffusion distillation},
  author={Sauer, Axel and Lorenz, Dominik and Blattmann, Andreas and Rombach, Robin},
  booktitle={European Conference on Computer Vision},
  pages={87--103},
  year={2024},
  organization={Springer}
}

@article{song2020denoising,
  title={Denoising diffusion implicit models},
  author={Song, Jiaming and Meng, Chenlin and Ermon, Stefano},
  journal={arXiv preprint arXiv:2010.02502},
  year={2020}
}

@article{song2023consistency,
  title={Consistency models},
  author={Song, Yang and Dhariwal, Prafulla and Chen, Mark and Sutskever, Ilya},
  journal={arXiv preprint arXiv:2303.01469},
  year={2023}
}

@article{song2023improved,
  title={Improved techniques for training consistency models},
  author={Song, Yang and Dhariwal, Prafulla},
  journal={arXiv preprint arXiv:2310.14189},
  year={2023}
}

@article{trippe2022diffusion,
  title={Diffusion probabilistic modeling of protein backbones in 3d for the motif-scaffolding problem},
  author={Trippe, Brian L and Yim, Jason and Tischer, Doug and Baker, David and Broderick, Tamara and Barzilay, Regina and Jaakkola, Tommi},
  journal={arXiv preprint arXiv:2206.04119},
  year={2022}
}

@article{wallner2023afsample,
  title={AFsample: improving multimer prediction with AlphaFold using massive sampling},
  author={Wallner, Bj{\"o}rn},
  journal={Bioinformatics},
  volume={39},
  number={9},
  pages={btad573},
  year={2023},
  publisher={Oxford University Press}
}

@article{wayment2024predicting,
  title={Predicting multiple conformations via sequence clustering and AlphaFold2},
  author={Wayment-Steele, Hannah K and Ojoawo, Adedolapo and Otten, Renee and Apitz, Julia M and Pitsawong, Warintra and H{\"o}mberger, Marc and Ovchinnikov, Sergey and Colwell, Lucy and Kern, Dorothee},
  journal={Nature},
  volume={625},
  number={7996},
  pages={832--839},
  year={2024},
  publisher={Nature Publishing Group UK London}
}

@article{wu2022high,
  title={High-resolution de novo structure prediction from primary sequence},
  author={Wu, Ruidong and Ding, Fan and Wang, Rui and Shen, Rui and Zhang, Xiwen and Luo, Shitong and Su, Chenpeng and Wu, Zuofan and Xie, Qi and Berger, Bonnie and others},
  journal={BioRxiv},
  pages={2022--07},
  year={2022},
  publisher={Cold Spring Harbor Laboratory}
}

@article{xu2019distance,
  title={Distance-based protein folding powered by deep learning},
  author={Xu, Jinbo},
  journal={Proceedings of the National Academy of Sciences},
  volume={116},
  number={34},
  pages={16856--16865},
  year={2019},
  publisher={National Academy of Sciences}
}

@article{yang2020improved,
  title={Improved protein structure prediction using predicted interresidue orientations},
  author={Yang, Jianyi and Anishchenko, Ivan and Park, Hahnbeom and Peng, Zhenling and Ovchinnikov, Sergey and Baker, David},
  journal={Proceedings of the National Academy of Sciences},
  volume={117},
  number={3},
  pages={1496--1503},
  year={2020},
  publisher={National Academy of Sciences}
}

@article{zhao2023unipc,
  title={Unipc: A unified predictor-corrector framework for fast sampling of diffusion models},
  author={Zhao, Wenliang and Bai, Lujia and Rao, Yongming and Zhou, Jie and Lu, Jiwen},
  journal={Advances in Neural Information Processing Systems},
  volume={36},
  pages={49842--49869},
  year={2023}
}
